\documentclass{article}

\usepackage{graphicx}
\usepackage{amsmath}
\usepackage{amssymb}
\usepackage{booktabs}
\usepackage{xcolor}
\usepackage[linesnumbered,ruled,vlined]{algorithm2e}
\usepackage{algpseudocode}
\usepackage{caption,subfigure}

\SetCommentSty{mycommfont}

\SetKwInput{KwInput}{Input}                % Set the Input
\SetKwInput{KwOutput}{Output}              % set the Output

\usepackage{balance}

\usepackage{pgfplots} % package used to implement the plot  
\pgfplotsset{compat=1.9}

\newcommand{\comment}[1]{}

\usepackage[pagebackref,breaklinks,colorlinks]{hyperref}

\usepackage[capitalize]{cleveref}

\newcommand{\selene}{\texttt{SELENE}}

\begin{document}

%%%%%%%%% TITLE - PLEASE UPDATE
\title{Self Semi Supervised Neural Architecture Search\\ for Semantic Segmentation}

\author{Loïc Pauletto$^{1,2}$, Massih-Reza Amini$^1$, Nicolas Winckler$^2$\\~\\
$^1$Universit\'e Grenoble Alpes \\{\tt\small firstname.lastname@univ-grenoble-alpes.fr}\\
\and 
$^2$ATOS \\{\tt\small firstname.lastname@atos.net}
% For a paper whose authors are all at the same institution,
% omit the following lines up until the closing ``}''.
% Additional authors and addresses can be added with ``\and'',
% just like the second author.
% To save space, use either the email address or home page, not both
}
\date{}
\maketitle

%%%%%%%%% ABSTRACT
\begin{abstract}
  In this paper, we propose a Neural Architecture Search strategy based on self supervision and semi-supervised learning for the task of semantic segmentation. Our approach builds an optimized neural network (NN) model for this task by jointly  solving a jigsaw pretext task discovered with self-supervised learning over unlabeled training data, and, exploiting the structure of the unlabeled data with semi-supervised learning. The search of the architecture of the NN  model is performed by dynamic routing using a gradient descent algorithm. Experiments on the Cityscapes and PASCAL VOC 2012 datasets demonstrate that the discovered neural network is more efficient than a state-of-the-art hand-crafted NN model with four times  less floating operations. 
\end{abstract}

%%%%%%%%% BODY TEXT
\section{Introduction}
\label{sec:intro}
Semantic segmentation entails in assigning a specific class to each pixel in an image with the overall aim of discovering objects. It is a key task in the field of computer vision, and has a wide range of applications, including autonomous driving \cite{badrinarayanan2017segnet}, medical research \cite{ronneberger2015u}, facial recognition \cite{muller2021convolutional} and person reidentification \cite{Wu2020reid}. In comparison to other computer vision tasks, the equivalent of this pixel-level label is a difficult and time-consuming effort.

The key challenges here are taking into account the context of objects inside images \cite{Mottaghi14}, and, learning with a small set of annotated data together with a large set of unlabeled data. To address these issues, different approaches have been proposed. For automatically taking into account the context, various approaches have been proposed under the self-supervised framework, which consists in exploiting the underlying data structure in order to gain supervision for an auxiliary task and learn a model by resolving both the auxiliary and the semantic segmentation problems simultaneously. For the problem of learning with partially labeled training data, or semi-supervised learning, existing approaches mostly assign pseudo-labels to unlabeled training data in order to augment the labeled training set using an auxiliary loss; under various perturbations such as images augmentations \cite{french2019semi}, features \cite{OualiHT20} or network \cite{KeWYRL19} perturbations; for enforcing the consistency of predictions. 

Although both self-supervised and semi-supervised approaches take advantage of data structure from different perspectives, they have never been studied together. Furthermore, hand-crafted NN architectures are used in the bulk of semantic segmentation research. Some recent works have proposed neural architecture search (NAS) for the design of a more flexible network that automatically adapt to the size of the input images \cite{LiSCLZWS20}. These NAS approaches, on the other hand, were developed with full supervision. 

In this work, we present a way to bridge between these three worlds. Our approach is based on self supervision and semi-supervised learning for semantic segmentation. It combines information from the labeled  training data with the resolution of a jigsaw pretext task discovered through self-supervised learning, and the search for regularities over the current model output over the labeled and unlabeled training data; in order to create an optimized neural network (NN) model for this task. Dynamic routing with a gradient descent technique is used to seek for the architecture of the NN model. 

We evaluate our approach with different settings where the model architecture is obtained by NAS when exploiting the context with either self-supervision or semi-supervised learning. For the latter, we employ two different semi-supervised techniques and show that learning with partially labeled data may not always lead to an efficient model. The performance of the architecture search is shown on known semantic segmentation benchmarks under different partition set-ups and are further compared to a state-of-the-art hand-crafted architecture.  We show that when searching for regularities on the outputs of the neural networks while concurrently addressing the pretext task, the proposed model achieves the best results compared to its other variants and a state-of-the-art hand-crafted NN with 4 times less floating-point operations (FLOPs).

The rest of the paper is organized as follows. Section~\ref{rw} presents an overview of the related work. Section~\ref{tech} provides details about the semi-supervised techniques used in our framework. Section~\ref{pf}, exhibits our approach, and finally, in Section~\ref{exper} we present experimental results obtained with our approach on Cityscapes and PASCAL VOC 2012 benchmarks. Finally, in Section~\ref{conclu} we analyze the study's findings and provide some suggestions for further research.

% %%%%%%%%%%%%%%%%%%%%%%%%%%%%%%%%%%%%% %
%           S E C T I O N   2           %
% %%%%%%%%%%%%%%%%%%%%%%%%%%%%%%%%%%%%% %

\section{Related Works}
\label{rw}
%This section provides an overview of works related to our study.

\subsection{Semantic segmentation}

Semantic segmentation consists in classifying each pixel of an image into a class, where each class represents an object or a portion of the image \cite{LiuDY19}. This task is part of the scene understanding concept, which is much more complex than image classification, as it requires apprehending the whole context of an image. To comprehend a scene, each piece of visual data must be assigned to an entity while taking into account the spatial information.    

Recent research on this topic has largely relied on Neural-Networks, as these models have been shown to outperform other techniques in scene analysis \cite{LongSD15}. 

Hand-crafted architectures, designed by experts in the field, are the most popular way to create specific NN models for semantic segmentation. In this category, a wide range of architectures requiring high computational resources exist, including U-Net \cite{ronneberger2015u}, Conv-Deconv \cite{FourureEFMT017} or FCN \cite{LongSD15}. However, \cite{YuWPGYS18} have shown that, by dissociating context information from the spatial information, it is possible to achieve highly efficient models with lighter architectures.  

Recently, \cite{ChenCZPZSAS18,NekrasovCS019}, have studied how Neural Architecture Search (NAS) can be applied to the decoder in order to  improve the performance for semantic segmentation. However, different from the proposed study, these techniques rely on a fully supervised learning framework.

\subsection{Self-supervised Learning}

%% Definition of Self-supervised Learning
Self-supervised learning entails automatically generating some sort of supervisory signal from the unlabeled data in order to achieve a task \cite{JingT21}. The auxiliary or pretext tasks extracted automatically are generally designed in such a way that solving them requires the learning of useful features. The majority of the new approaches are built on the concept of contrast (i.e. contrastive learning), with a learning procedure that uses network representations of images to learn strong and useful features \cite{He0WXG20, CaronMMGBJ20, GrillSATRBDPGAP20}. In addition, there are other approaches where the goal is to colorize the image \cite{ZhangIE16}; or to fill in missing parts of the image \cite{PathakKDDE16} or even to predict the direction of rotation of an image \cite{GidarisSK18}.

For the task of classification, some of the first papers have proposed full image-based methods, that is, without patches \cite{GidarisSK18,DosovitskiyFSRB16}. Different from that, more recent approaches have proposed patch-based methods. These methods were devised to estimate the relative positions of two non-overlapping patches in a 3$\times$3 grid \cite{DoerschGE15}. Other approaches, generalizing this concept,  involve using the entire grid and attempting to solve a jigsaw  by anticipating the permutation used to shuffle it \cite{NorooziF16}.

Recently, some research have proposed NAS in a self-supervision setting \cite{LiuDHGYX20}, in which an architecture is discovered  using a pretext task and then transferred to a supervised learning context.
%None of these strategies, however, were based on NAS. 
In this paper, we propose determining the NN architecture for semantic segmentation utilizing two forms of context information: solving a pretext task identified by self-supervision and exploiting the data structure through semi-supervised learning. 

\subsection{Semi-supervised Learning}

%% Definition of Semi-Supervised Learning
Semi-supervised learning refers to the process of learning a prediction function from both labeled and unlabeled training examples. In this situation, labeled examples are supposed to be few, whereas unlabeled training data are available in abundance. Unlabeled samples include useful information about the prediction problem at hand, which are exploited jointly with the labeled training examples to produce a more efficient prediction function than if just the latter were used for learning \cite{Chap06,AminiU15}. 

Semi-supervised learning has a long history, and has been studied in a variety of domains and tasks \cite{Chap06, EngelenH20, zhu2005semi}. More recently, semi-supervised learning has gained popularity in the deep learning community, and has shown, in some cases, comparable results to the state-of-the-art of purely supervised methods \cite{LaineA17, MiyatoMKI19, TarvainenV17}. Most popular approaches  to deep semi-supervised learning \cite{Ouali2020AnOO} include for example  generative models, consistency regularization, entropy minimization, or proxy-label methods.

Semi-supervised learning methods, exploiting generative model based approaches, have been performed using e.g. variational autoencoders  \cite{KingmaMRW14,AbbasnejadDH17} or generative adversarial networks \cite{Odena16a, DaiYYCS17}.  These approaches aim to combine information from the joint distribution $p(x,y)$ of labeled data and the density distribution $p(x)$ of unlabeled data.

Neural network approaches based on consistency regularization add an auxiliary loss computed over unlabeled samples. This loss computes the divergence between predictions made on unlabeled perturbed data points. Among the first works using this approach, the use of Ladder Networks with additional noisy encoder, encoder and decoder was proposed\cite{RasmusBHVR15}, where they used a consistency regularization loss to denoise representations at each layer. Better results were then obtained by smoothing these predictions, e.g. using Temporal Ensembling \cite{LaineA17},  Mean Teacher \cite{TarvainenV17}, or Virtual Adversarial Training \cite{MiyatoMKI19}. More recent results were improved with fast-SWA \cite{AthiwaratkunFIW19}, using cyclic learning rates and measuring discrepancy with a set of predictions from multiple checkpoints. Similar to consistency training, which enforces consistent predictions under perturbations, pushing away the decision boundary toward low density regions, entropy minimization accomplish the same goal by increasing the prediction confidence on unlabeled data\cite{OlssonTPS21}.

Proxy-label methods iteratively assign pseudo-labels to high-confident unlabeled examples and include these pseudo-labeled examples in the learning process. These pseudo labels can either be produced by Self-training \cite{XieLHL20}, that is by the model itself, or by multi-view training, using for example the Co-training framework \cite{QiaoSZWY18}. Finally, hybrid, or \textit{holistic} methods such as MixMatch\cite{BerthelotCGPOR19}, ReMixMatch\cite{remixmatch}, or FixMatch\cite{SohnBCZZRCKL20} unified different approaches, combining for example consistency regularization and pseudo-labeling as in FixMatch. More recently, another example of \textit{holistic} approach, using Cross Pseudo supervision\cite{ChenYZ021} on the task of semantic segmentation, have shown state-of-the-art results.
% }

The majority of the cited methods employ a fixed architecture as their backbone; however, our approach uses information from both partially labeled data and context information, extracted by self-supervised learning, to automatically explore a specific and more flexible architecture for semantic segmentation. Moreover, the obtained model does not require to be retrained, as both, the architecture and its parameters, are learned simultaneously.

%%%%%%%%%%%%%%%%%%%%%%
%%% SECTION 3
%%%%%%%%%%%%%%%%%%%%%%
\section{Framework and baselines} % Discussions
\label{tech}
We assume that we have a labeled training set $\mathcal{D}_\ell=~(x_l,y_l)_{1\leq l\leq m}$ of size $m$, and a possibly much larger set of unlabeled training examples $\mathcal{D}_u=(x_u)_{m+1\leq u\leq m+n}$ of size $n$.  We further consider that $\theta$ represents the set of all network weights. 

In our setting, we consider the jigsaw solving pretext task as the self-supervised method. The main motivation here is that performing well on jigsaw puzzles necessitates a thorough comprehension of scenes and objects \cite{NorooziF16} which is also closely related to semantic segmentation. On the other hand, there is no apparent consensus in the literature on which semi-supervised approach is the most efficient for semantic segmentation. Here we considered the Mean-Teacher and the Co-Teaching approaches, which have been increasingly popular in recent years. Our goal is to investigate their efficacy in the context of semantic segmentation using neural architecture search. 
Depending on the self-supervised and semi-supervised $method$, we define $\mathcal{L}_u^{method}$ as the unsupervised loss related to the considered approach (i.e. $method\in\{ssl,ssup\}$). In all scenarios, the supervised loss $\mathcal{L}_s$ is based on an individual loss $\ell_d$ defined as the cross-entropy or the OHEM \cite{ShrivastavaGG16} loss, depending on the dataset.
%based on the individual cross-entropy loss $\ell_{ce}(.)$.
%is the average empirical cross-entropy estimated over the labeled training set. We will go through these approaches in further detail in the next sections. 

\subsection{Self-supervised regularization}
\label{sec:ssl}
\begin{figure}[t!]
  \centering
  %\fbox{\rule{0pt}{0.5in} \rule{0.9\linewidth}{0pt}}
  \includegraphics[width=0.8\linewidth]{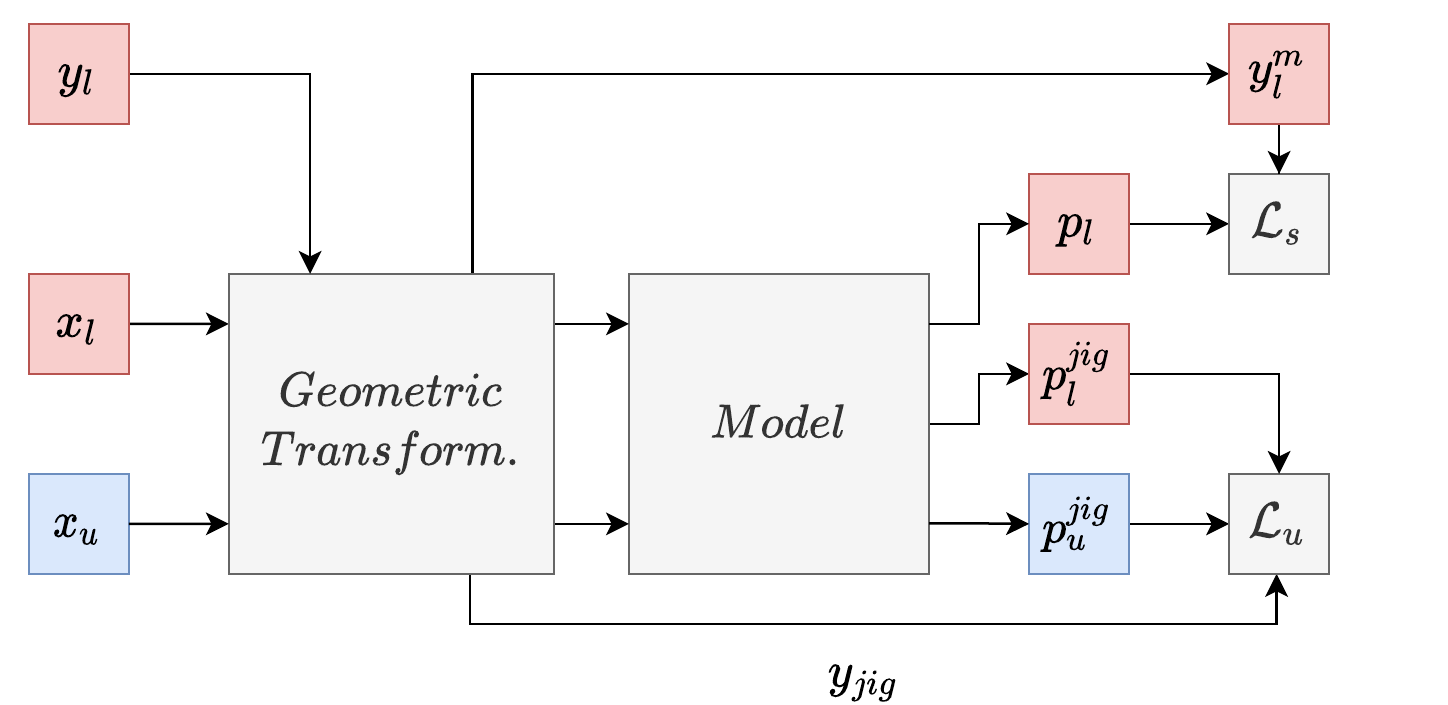}
   \caption{Illustration of the self-supervised learning strategy, $x_u$ is an unlabeled sample and $(x_l,y_l)$ is a labeled training example. The label $y_l^{jig}$ is the transformed version of $y_l$, where $jig$ is the id of the applied jigsaw permutation.  $(p_l)_{1\leq l\leq m}$ are predictions for the supervised semantic segmentation task, $p_l^{jig}$ and $p_u^{jig}$ are the prediction for the pretext task.}
   \label{fig:ssl}
\end{figure}

For the self-supervised learning method, a geometric transformation is applied to the inputs for the pretext task, and a label is generated. For each labeled training example, $(x_l,y_l)\in\mathcal{D}_\ell$ this transformation on the input $x_l$ acts as an augmentation and the same transformation is applied on $y_l$. For an unlabeled example $x_{u}\in\mathcal{D}_u$, the label produced by the transformation, $y_{u}$, is  used as the ground truth with respect to the pretext task. As proposed in \cite{NorooziF16}, the key idea of the jigsaw pretext task is to learn visual representations for puzzles solving. In practice, this task consists in cutting images in 9 patches from a grid of $3\times3$. The patches are then mixed using specified random permutations, and the network is trained to predict the permutation in question in order to solve the problem. The suggested self-supervised learning strategy is depicted in Figure \ref{fig:ssl}. Along with the supervised semantic segmentation problem, one or more distinct pretext tasks can be considered in this framework.
In addition, unlike other state-of-the-art approaches, just one network is employed, and the perturbation is applied to the input via geometric transformation. 

% \begin{figure}[h!]
%   \centering
%   \fbox{\rule{0pt}{2in} \rule{0.9\linewidth}{0pt}}
%   %\includegraphics[width=0.8\linewidth]{egfigure.eps}
%   \label{fig:onecol}
% \end{figure}
%\subsubsection{Inpainting}
%Based on context encoder\cite{} works, here the key idea is to fill missing part in the image. The self-supervised loss function is defined as $L_2$ like follows:
%\begin{equation}
%    L_{Inpainting} = \norm{M \odot (x - t f_\theta((1-M) \odot x))}_2^2
%\end{equation}
%where $f_\theta$ represent the model with $\tehta$ parameters, $M$ is binary mask and $\odot$ the element-wise product.

%\subsubsection{Denoising auto-encoders}
%In the continuity of the previous task, the task of denoising images has been used for a long time to learn robust features\cite{}. In this task the goal is to recover an image from a noisy version. The key idea of this task is inspired by the fact that humans can recognize object even in noisy images, indicating that visual features can be extracted separated from noise. The self-supervised is defined as follows:
%\begin{equation}
%    L_{noise} = \norm{x - f_\theta(z \odot x)}_2^2
%\end{equation}

%\subsubsection{Jigsaw solving}
In our experiments, we followed a similar approach to \cite{BeyerZOK19}, by training the network in a multitask manner, where a supervised loss ($\mathcal{L}_s$) is minimized along a self-supervised loss ($\mathcal{L}_u^{ssl}$) that acts as a regularizer.
In this case, for a given permutation $jig\in\{1,\ldots,k\}$ where $k$ is the total number of considered permutations; the problem of jigsaw solving can be formulated as a classification task using the produced pretext labels for both labeled  $(y_l^{jig})_{1\leq l\leq m}$ and unlabeled training samples $(y_u^{jig})_{1\leq u\leq n}$. The self-supervised loss function is in this case, the average cross-entropy loss:
{\small
\begin{equation*}
\mathcal{L}_u^{ssl} = \frac{1}{k}\sum_{jig=1}^{k}\left( \frac{1}{m}\sum_{\mathcal{D}_\ell}\ell_{ce}(p_l^{jig}, y_l^{jig})+\frac{1}{n}\sum_{\mathcal{D}_u}\ell_{ce}(p_u^{jig}, y_u^{jig})\right)
\label{loss_self}
\end{equation*}
}
where $\ell_{ce}(.)$ is the individual cross-entropy loss function. In all of ours experiments, we used a fixed set of $k=100$ permutations, as in  \cite{NorooziF16}. For the supervised loss, we consider the average cross-entropy for the supervised semantic segmentation task over the labeled training set:
\begin{equation}
   \mathcal{L}_s = \frac{1}{m} \sum_{\mathcal{D}_l} \ell_{d}(p_l, y_l) ,
    \label{ssl_loss_sup}
\end{equation}

%\subsubsection{Rotation}
%Here, as proposed in \cite{}, the idea of the rotation pretext task is to rotate an input $x$, and then to predict the rotation angle applied to this input. This method has been mainly used in semi-supervised classification and have shown improvement over the baseline. The rotation angle is randomly applied to $x$ among four possibility $\{0^{\circ}, 90^{\circ}, 180^{\circ}, 270^{\circ}\}$. In this case, the problem is a simple 4-class classification, thus the self-supervised loss is the standard cross-entropy.

%\subsection{Semi-supervised methods}
%In this section, we will introduce the various semi-supervised methods that we study in this work.
%\textcolor{blue}{We decided to study the following methods, because of their simplicity and efficiency in the state of the art.}

\subsection{Semi-supervised Mean Teacher method}
\label{sec:MT}
The mean-teacher method was developed for semi-supervised classification \cite{TarvainenV17} under the self-training paradigm \cite{Amini:2008}. The approach has lately been adapted to semi-supervised semantic segmentation \cite{french2019semi}.

\begin{figure}[b!]
  \centering
  %\fbox{\rule{0pt}{0.5in} \rule{0.9\linewidth}{0pt}}
  \includegraphics[width=0.7\linewidth]{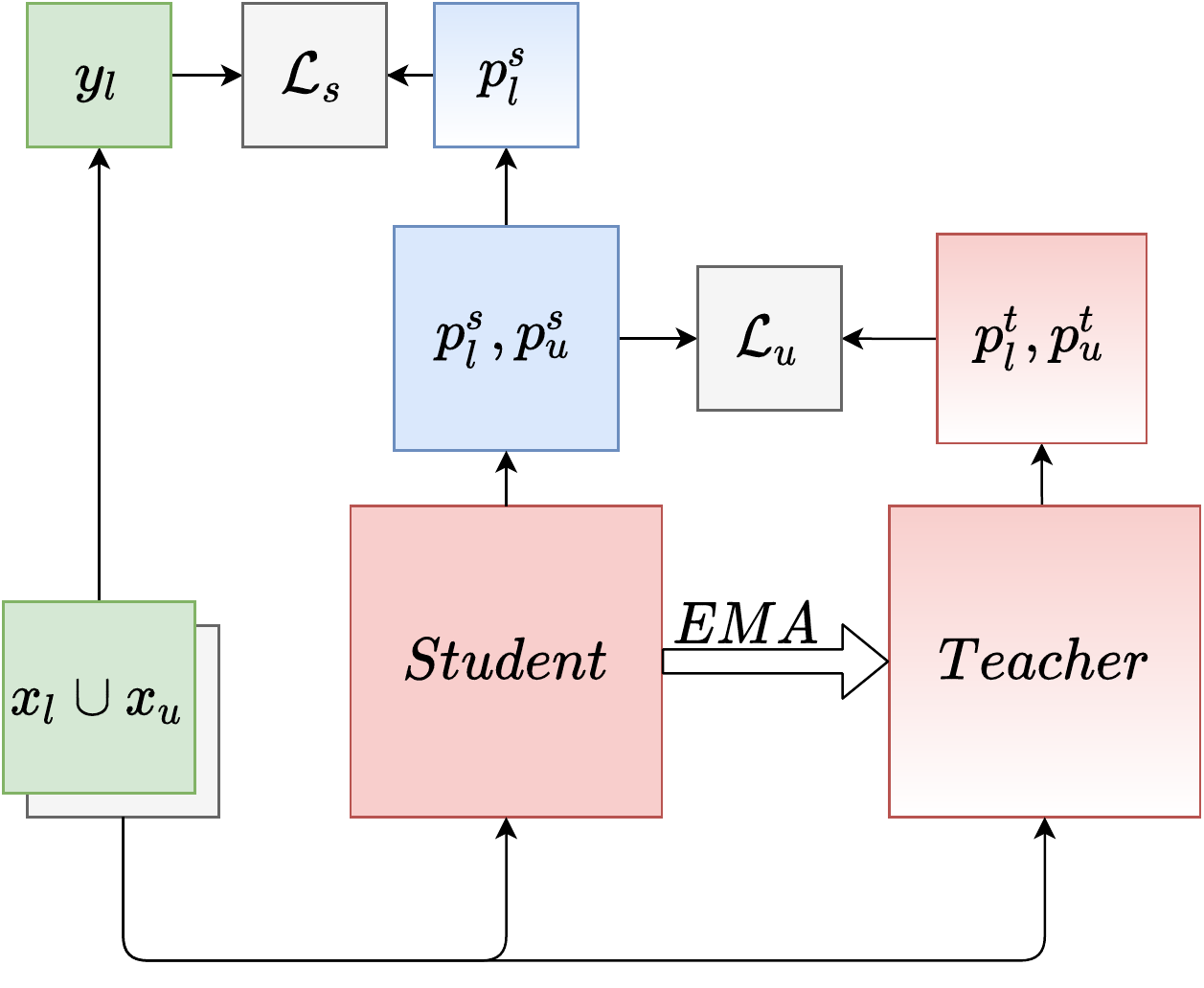}
   \caption{Illustration of mean teacher process. $p^s_l,p^s_u$ and $p^t_l,p^t_u$ are predictions for labeled and unlabeled samples, by the student and the teacher, respectively. The EMA arrow stand for exponential moving average of weights from the student to the teacher.} 
   \label{fig:mt}
\end{figure}
This approach is based on consistency regularization that constraints a model to have the same output for a given input. The underlying model is made of two neural networks of the same structure, one denoted as \textit{student}, $f(\theta)$, and the other called  \textit{teacher}, $f(\overline{\theta})$), where $\overline{\theta}$ denote the exponential moving average (i.e. EMA) of the parameters of the student, and are  iteratively computed as:
\begin{equation}
    \overline{\theta}_{t+1} = \alpha\overline{\theta}_{t} + (1-\alpha){\theta}_{t},
    \label{ema_formula}
\end{equation}
with $\alpha$ a hyperparameter used to control the dependency between the two networks. In this method, only the \textit{student} is trained using the labeled training set and the predictions of the teacher. Let $(p^s_l)_{1\leq l\leq m}$ and $(p^t_l)_{1\leq l\leq m}$ be the outputs predicted by the \textit{student} and \textit{teacher} networks on labeled examples; and, $(p^s_u)_{1\leq u\leq n}$ and $(p^t_u)_{1\leq u\leq n}$ predictions of both networks on unlabeled samples. The consistency regularization is achieved by restricting the distribution outputs of both networks to be as close to each other as possible on labeled and unlabeled samples provided as inputs to both models; and that by minimizing the following unlabeled loss by the student:
{\small
\begin{equation}
\mathcal{L}_u^{ssup-mt} =  \frac{1}{m}\sum_{\mathcal{D}_\ell}\ell_{MSE}(p_l^t, p_l^s) + \frac{1}{n}\sum_{\mathcal{D}_u}\ell_{MSE}(p_u^t, p_u^s)
    \label{mt_loss}
\end{equation}
}
where $\ell_{MSE}$ is the Mean-Square Error summed up over all pixels and classes. The supervised loss of the student is based on the \textit{teacher} outputs and the ground truth of the labeled training examples:
\begin{equation}
   \mathcal{L}_s = \frac{1}{m} \sum_{\mathcal{D}_l} \ell_{d}(p_l^t, y_l) ,
    \label{mt_loss_sup}
\end{equation}

Figure \ref{fig:mt} illustrates this strategy. At the beginning, both  networks have the same initial weights. At each iteration, the parameters of the student are first updated by minimizing both the supervised and unsupervised losses; then the parameters of the teacher are updated by EMA \eqref{ema_formula}. Following \cite{french2019semi}, we fixed $\alpha$ to $0.99$  in  our experiments.

%In equation \ref{mt_loss}, $\hat{y}_l^{teacher}$ and $\hat{y}_u^{teacher}$ represent the output probability distribution of the teacher for the labeled and unlabeled samples, respectively. These notations are the same for the student outputs.

\begin{figure}[b!]
  \centering
  %\fbox{\rule{0pt}{0.5in} \rule{0.9\linewidth}{0pt}}
  \includegraphics[width=0.55\linewidth]{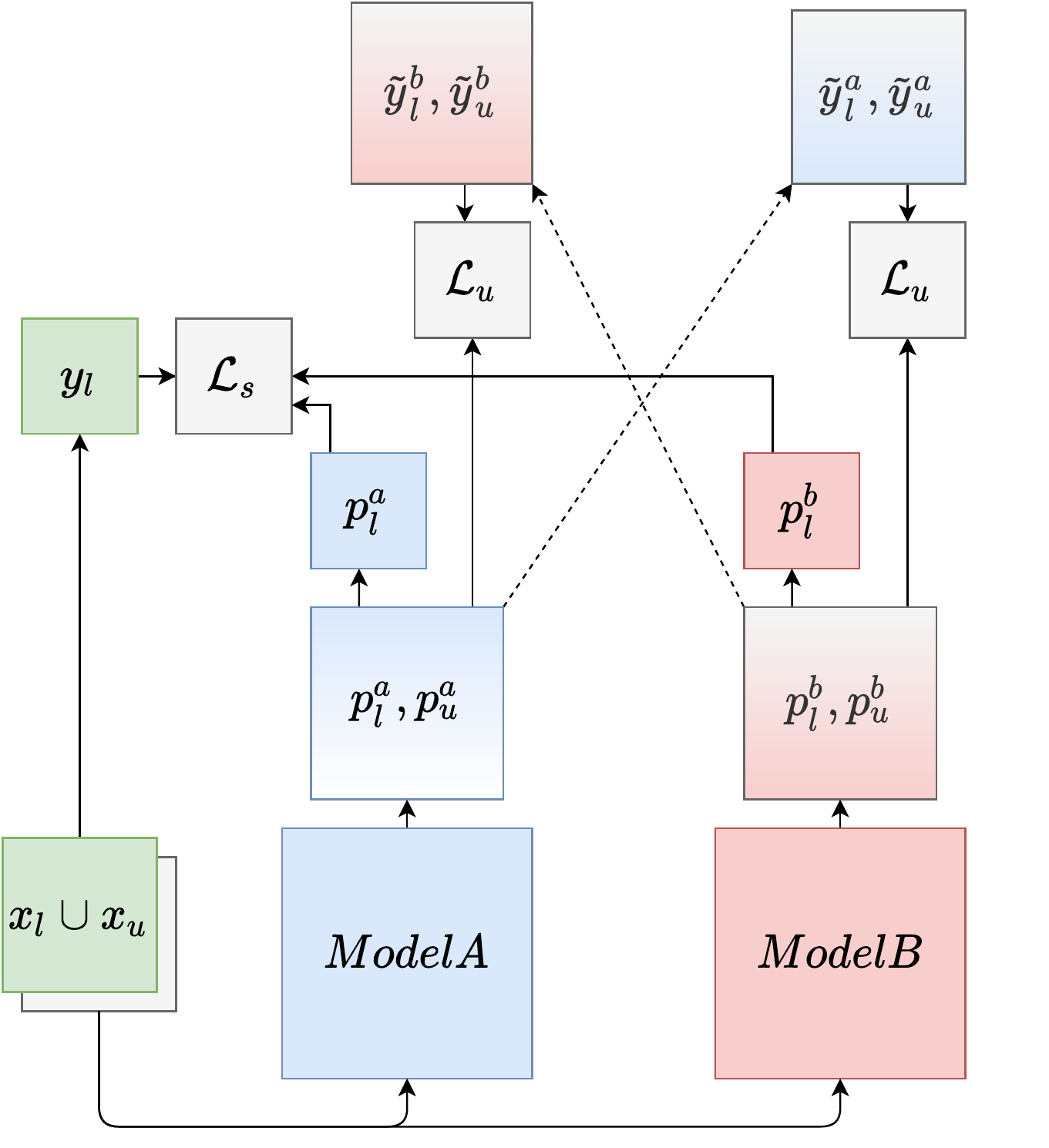}
   \caption{Illustration of the co-teaching workflow, where $x_u$ is an unlabeled sample, $(x_l,y_l)$ is a pair of labeled samples. The terms $p_l^a, p_u^a$ and $p_l^b, p_u^b$ are predictions for labeled and unlabeled samples of the Model A and B respectively. $\tilde{y}_l^a, \tilde{y}_u^a, \tilde{y}_l^b, \tilde{y}_u^b$ are the associated pseudo-labels obtained with one-hot encoding.}
   \label{fig:cot}
\end{figure}

\subsection{Semi-supervised co-teaching method}
\label{sec:CT}
This method is based on self-training \cite{Amini:2008} and it is well-known in the deep learning community, and it has been widely used in different problems, including semantic segmentation \cite{ChenYZ021}. Similarly to the mean teacher approach, co-teaching uses two networks of the same architecture (e.g. $f(\theta_a)$, $f(\theta_b)$). However, differently from the previous approach, here the two networks update their weights and are independently trained. The main idea is that each network can learn from the other one. Formally, given an unlabeled input $x_u$, each network, i.e. $f(\theta_a)$ and $f(\theta_b)$, predicts an output ${p}_{u}^a$ and ${p}_{u}^b$. Then by one-hot encoding, these outputs are transformed to pseudo-labels; $\tilde{y}_u^a$ and $\tilde{y}_u^b$; which serve as ground truth for the other network, and the definition of a cross-pseudo supervision loss over the unlabeled data:
\begin{equation}
   \mathcal{L}_u^{cps} = \frac{1}{n} \sum_{\mathcal{D}_u} (\ell_{ce}(p_u^a, \tilde{y}_u^b) + \ell_{ce}(p_u^b, \tilde{y}_u^a)),
    \label{cot_loss}
\end{equation}

Similarly to $\mathcal{L}_u^{cps}$, a cross-pseudo supervision loss $\mathcal{L}_l^{cps}$ can be defined on the examples in $\mathcal{D}_\ell$ and the unsupervised loss is defined as the sum of these two losses:
\begin{equation}
   \mathcal{L}_u^{ssup-ct} = \mathcal{L}_u^{cps} + \mathcal{L}_l^{cps}
    \label{cot_loss_cps}
\end{equation}
The supervised loss that is used for the training of both networks is defined as:
\begin{equation}
   \mathcal{L}_s = \frac{1}{m} \sum_{\mathcal{D}_l} (\ell_d(p_l^a, y_l) + \ell_d(p_l^b, y_l)),
    \label{cot_loss_sup}
\end{equation}
Figure \ref{fig:cot} illustrates the whole process of co-teaching. In our setup both models, $f(\theta_a)$ and $f(\theta_b)$ are dynamic, and their weights are found independently one from the other. 
%%%%%%%%%%%%%%%%%%%%%%
%%% SECTION 4
%%%%%%%%%%%%%%%%%%%%%%
\section{\selene{}: Self sEmi-supervised LEarning with NEural Architecture Search } %Wrapper
\label{pf}

This section presents our self-supervised semi-supervised learning approach for semantic segmentation based on Neural Architecture Search, that we call \selene{}.

\subsection{\selene{} learning scheme}
Our learning problem for semantic segmentation is to jointly find an optimized architecture of a neural network and its parameters, $\theta$, that minimize the weighted sum of a supervised loss, $\mathcal{L}_{s}$; and two unsupervised losses defined from self-supervision, $\mathcal{L}_u^{ssl}$, and semi-supervised learning $\mathcal{L}_u^{ssup}$:
\begin{equation}
\label{eq:totalloss}
\mathcal{L}_{total}= \lambda_0 \mathcal{L}_{s}+\lambda_1 \mathcal{L}_u^{ssl}+\lambda_2 \mathcal{L}_u^{ssup}
\end{equation}
This idea is based on the premise that for the difficult task of scene analysis, which is highly dependent on context, it is required to use a specific model to leverage low and high-level information from the data. The low-level information here are the associations between the pixels of an image and their classes, which are present in the labeled training set. The high-level information is recovered from the unlabeled training data by first resolving the auxiliary jigsaw pretext task using self-supervised learning and then using semi-supervised learning to exploit the structure of the unlabeled data. The pseudocode of \selene{} is depicted in Algorithm \ref{alg:cap}. The neural architecture search is performed using the Dynamic Routing  approach presented in the next section. First, the routing path is arbitrarily set, and model weights are initialized using imageNet pre-trained weights and Kaiminig initialization (Section \ref{implem_detail}). At each epoch $e$, defined as the largest size between the labeled and the unlabeled training sets, two batches of data;  $\mathfrak{B}_\ell$ and $\mathfrak{B}_u$; are randomly extracted from these two sets.  Unlabeled losses; $\mathcal{L}_u^{ssl}$ and $\mathcal{L}_u^{ssup}$; are then set using the self-supervised regularization and the semi-supervised method with the current neural network's predictions as presented in Section \ref{tech}. A new routing path and weights are then found by minimizing the total loss, $\mathcal{L}_{total}$ \eqref{eq:totalloss}.

\begin{algorithm}
    \DontPrintSemicolon
    \SetAlgoLined
    \SetKwInOut{Input}{Input}
    \SetKwInOut{Output}{Output}
    \Input{$\mathcal{D}_u$, $\mathcal{D}_\ell$, $E$: \textit{epochs}, $M$: \textit{method}, $\lambda_0,\lambda_1,\lambda_2$: \textit{losses weights}}
    $N \gets \max(|\mathcal{D}_u|,|\mathcal{D}_\ell|)$ \;
    \tcc{The Dynamic routing structure (Sect. \ref{DR_ssl}) is used to set up the initial architecture.}
    $f_{\theta} \gets \texttt{DR\_struture()}$ \;
    \tcc{Initialization of the weights}
    $f_{\theta_{0}} \gets$ \texttt{init()} \; %\tcp*{\small{ Random initialization of the weights}}
    %\tcc{Now this is a While loop}
    \For{$e \in \texttt{range(0,E)}$} 
    {
        \For{$i \in \texttt{range(0, N)}$} 
        {
           $t \gets i + e$\;
           \tcc{Get batch of labeled samples}
           $\mathfrak{B}_\ell \gets \mathcal{D}_\ell$\;
           \tcc{Get batch of unlabeled samples}
           $\mathfrak{B}_u \gets \mathcal{D}_u$ \;
          % \If{\texttt{use}\_$\selene{}^+$}{
           $\mathcal{L}_u^{ssl}, \mathcal{L}_u^{ssup} \gets M(\mathfrak{B}_l, \mathfrak{B}_u, f_{\theta_t}) $\;
        $\mathcal{L}_{total} \gets  \lambda_0 \mathcal{L}_{s} +  \lambda_{1} \mathcal{L}_u^{ssl} + \lambda_{2} \mathcal{L}_u^{ssup}$\;
          % }
         %  \ElseIf{\texttt{use}\_$\selene{}^\dagger$}{
         %%  $\mathcal{L}_s, \mathcal{L}^{M} \gets M(\mathfrak{B}_l, \mathfrak{B}_u, f_{\theta_t}) $\\
       %%  $\mathcal{L}_{total} \gets  \lambda_0 \mathcal{L}_{s} + \lambda_1 \mathcal{L}^{M}$
          % }
          % \Else{
          %%%%  $\mathcal{L}_{total} \gets  \lambda_0 \mathcal{L}_{s} + \lambda_1 \mathcal{L}^{ssl}$
         %  }
           $f_{\theta_{t+1}} \gets$\texttt{DR\_optimize$(f_{\theta_t}, \mathcal{L}_{total})$\\
        }   
        }
    }
    \Output{$f_{\theta^\star}$: \textit{Network with trained weights after $E$ epochs}}
    
    \caption{Pseudo-code of \selene{}}\label{alg:cap}
\end{algorithm}

\subsection{Architecture Search with Dynamic Routing}
\label{DR_ssl}

Dynamic networks have exhibited superiority in network capacity and greater performance with budgeted resource use, by fitting the model's architecture to the input data.  Among different approaches, dynamic routing \cite{LiSCLZWS20}, on which we base our routing algorithm, has the advantage of allowing the transfer of weights from a prior training, that has become more essential in terms of time savings. 
\begin{figure*}[h] % mettre en grand ?
  \centering
  %\fbox{\rule{0pt}{0.5in} \rule{0.9\linewidth}{0pt}}
  \includegraphics[scale=0.4]{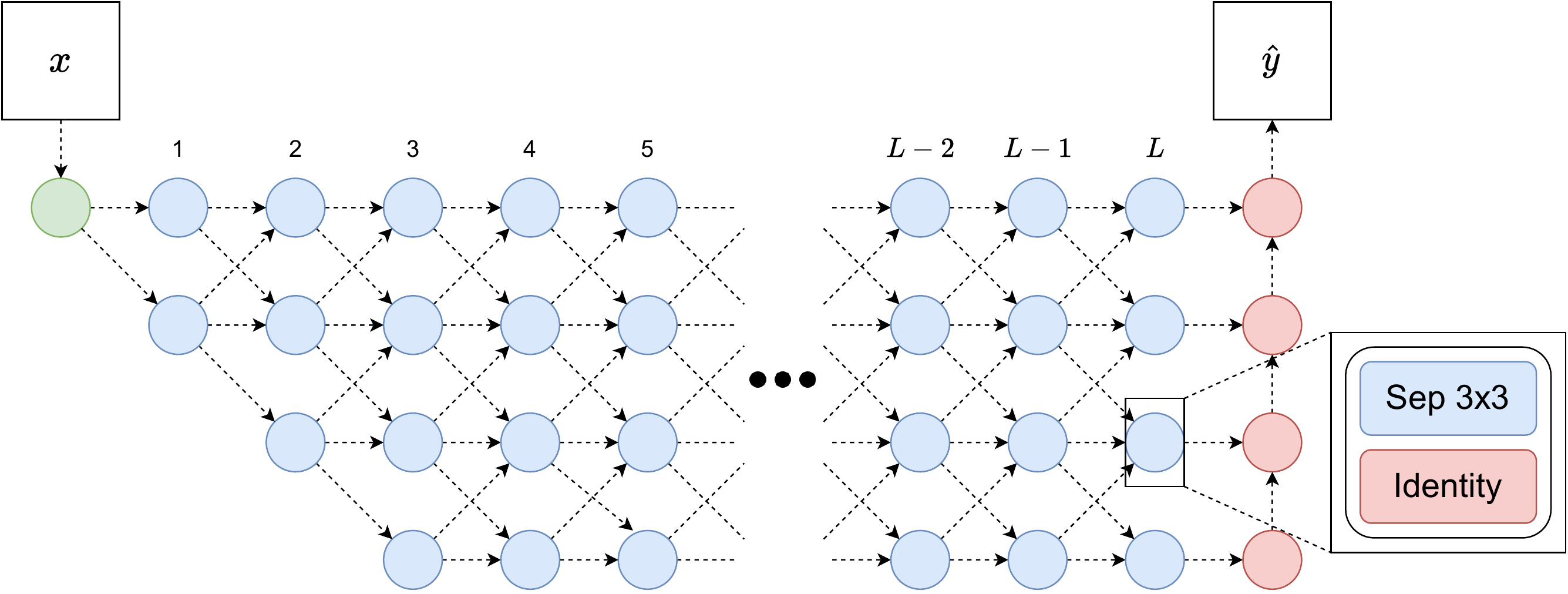}
   \caption{Illustration of the dynamic routing structure. This structure is composed of a part called 'STEM', the green cell, and a decoder, the red cells. The blue cells part is the encoder, each cell contains two operations a \textit{Separable convolution 3x3} and an \textit{Identity}. }
   \label{fig:dn}
\end{figure*} 

In our approach, the routing space (or structure) noted $f_\theta$, is defined as a 4-level network with $L$ layers composed of cells (Figure \ref{fig:dn}). Each level in this structure represents a stride rate, where the size of the input is successively reduced by descending in the network. The strides rates are thus $1/4$ for the highest level and $1/32$ for the deepest one. Depending on whether the level is greater or lower, the image ratio is then lowered or raised by $2$.  The path through the levels is performed by a convolution with a kernel size of $1$, while the size is reduced, the convolution increases the number of feature map by 2. 

For the initialization of the dynamic structure, we set a fixed 3-layer block \textit{'STEM'} (green cell in Figure \ref{fig:dn}), used to reduce the resolution of the input to $1/4$. Note that in this block, separated convolution are used. At the end, we find a simple decoder (red cells in Figure \ref{fig:dn}), which go from bottom to top of the levels. This decoder is just a composition of convolution and upsampling operations in order to add the features of each level. Once added, the features go through a last classification layer. Concerning the cells aspect (blue cells in Figure \ref{fig:dn}), each cell is made up of two operations, which are here a separable convolution with a kernel of 3 and an identity operation. The choice of the identity operation will result in the creation of a skip-connection. The choice of the path in this structure, the choice of the cell operations and the parameters of each operation are all optimized using the gradient descent algorithm. 

%%%%%%%%%%%%%%%%%%%%%%
%%% SECTION 5
%%%%%%%%%%%%%%%%%%%%%%

\section{Experiments}
\label{exper}

We ran a series of experiments to study how the combination of self-supervised and semi-supervised learning with NAS can help to take advantage of unlabeled training data to construct an efficient NN for semantic segmentation.

\subsection{Datasets}
% We perform experiments on the following datasets.
\paragraph{Pascal VOC 2012}
The Pascal VOC dataset \cite{EveringhamGWWZ10}, which contains 20 object classes and one background category, is a widely used dataset in object semantic segmentation. The original dataset contains almost 13000 images, including 1464 images for training, 1449 for validation, and 1456 for testing as standard splits. We employ the augmented version provided by \cite{HariharanABMM11} as a standard base for our work, raising the total number of usable images for training to 10582.
\paragraph{Cityscapes}
The Cityscapes dataset \cite{CordtsORREBFRS16} is frequently utilized, mostly in the context of analyzing urban scenes. The collection contains 5000 finely annotated images, each with a per-pixel label from one of 19 semantic classes. There are 2975 images for training, 500 for validation, and 1525 for testing in the splits provided, with each image having a resolution of 1024x2048. Following other studies, we solely use the training and validation sets in our experiments. There is also a part known as \textit{coarse}, which contains 20,000 images with coarse annotations; not used in our experiments.
% that we did not use in our experiments.

\subsection{Experimental setup}
\label{implem_detail}
\paragraph{}For both datasets, similar data augmentation to  \cite{ChenYZ021,LiSCLZWS20} have been used. Random scaling, followed by random horizontal flipping, and random square cropping have been applied as augmentation. 
The scaling factor was taking values  in $\{0.5, 0.75, 1, 1.25, 1.5, 2.0\}$, and the crop size ranges from $800$ for Cityscapes and $512$ for Pascal VOC, with padding, and an ignored value if necessary. We set the hyperparameter of the supervised loss in \eqref{eq:totalloss}; $\lambda_0 = 1$ in all of our experiments. 

To investigate the effects of semi-supervised and self-supervised settings on the learning of parameters, we respectively set the corresponding hyperparameters; $\lambda_1$ and $\lambda_2$; in \eqref{eq:totalloss} to $0$. These scenarios will be presented in the next section. We deploy an extra classifier for the pretext task in the self-supervised experiments. Accordingly, the dynamic routing output is taken in the other direction (from top to bottom in Figure \ref{fig:dn}). Then, before the fully connected layer, we apply global average pooling to the features.

Concerning technical aspects, \selene{} is implemented using PyTorch\cite{pytorch} library, and trained using Nvidia GPUs. The encoder of the network (i.e. blue dots in Figure \ref{fig:dn}) is initialized by the ImageNet pre-trained weights, provided by \cite{LiSCLZWS20}, while the others weights (i.e. red and green dots in Figure \ref{fig:dn}) are initialized using Kaiming initialization \cite{HeZRS15}. For parameter updates, we used a standard mini-batch SGD with momentum of 0.9, with an initial learning rate of 0.02. In addition, we adopt a polynomial learning rate decay with a power of 0.9. The training batch size is $8$ for Cityscapes and $16$ for Pascal VOC. Concerning the dynamic routing structure, we follow \cite{LiSCLZWS20} and set $L=16$. 

Finally, results are reported on the full validation set for each dataset (500 for Cityscapes, 1449 for Pascal VOC) using the standard mean Intersection-over-Union (mIoU) metric \cite{RezatofighiTGS019}. In all of our experiments, we use single scale testing with no augmentation.

%This decoder is different from the one used for segmentation, because we used it in the opposite way (from top to bottom if fig. \ref{fig:dn}). Then, we apply global average pooling to the features, before the classifier.

\subsection{Experimental results}
In this section, we present the experimental results obtained by \selene{} under various settings. For all the experiments, we use the same partitions as proposed in \cite{ChenYZ021}. 

\subsubsection{Self-supervised regularization}

We begin by examining the obtained gain  by performing the jigsaw pretext problem discovered by self-supervised learning  (Section \ref{sec:ssl}) simultaneously with the pixel classification task for semantic segmentation. In this scenario, the corresponding hyperparameter of self-supervised learning in \eqref{eq:totalloss}; $\lambda_1$ was set to $0.1$; and; we disabled the effect of semi-supervised learning  by setting the hyperparameter $\lambda_2$ to $0$. The corresponding model is denoted by  \selene$_{\lambda_2=0}$ and it is compared to the fully supervised setting, in which no self-supervised nor semi-supervised learning is utilized. In the following, \selene$_{\lambda_1=\lambda_2=0}$, stands for the fully supervised model. Table \ref{tab:ssl} summarizes results obtained for different fraction of the labeled training set on PASCAL VOC.  The highest performance rates are indicated in boldface. It turns out that the pretext task effectively adds information to semantic segmentation, although the benefits are limited. This could be due to the fact that images are cut using a $3\times 3$ grid for puzzle solving, and the resolution of the jigsaw problem may introduce noise into the pixel classification, particularly where puzzle pieces are cut.

\setlength{\tabcolsep}{5pt}
\renewcommand{\arraystretch}{1}
\begin{table}[h]
    \centering
    \begin{tabular}{ccccc}
% \cline{2-5}
 \cmidrule[\heavyrulewidth](lr){2-5}
 \multicolumn{1}{l}{}  & \multicolumn{4}{c}{Fraction of the labeled training set} \\
       \cmidrule(lr){2-5}
    \multicolumn{1}{l}{}  & $\frac{1}{16}$  & $\frac{1}{8}$ & $\frac{1}{4}$  & $\frac{1}{2}$  \\
        \toprule
        % &$\frac{1}{16}$ (662) & $\frac{1}{8}$ (1323)& $\frac{1}{4}$ (2646) & $\frac{1}{4}$ (5291) \\
        % \hline
       \scriptsize \selene$_{\lambda_1=\lambda_2=0}$ & 53.33 & 59.45 & 65.21 & 69.02 \\
       % \hline
        \selene$_{\lambda_2=0}$ & \textbf{53.80} & \textbf{60.33} & \textbf{65.25} & \textbf{69.27} \\
        \bottomrule
        %\selene{}$\texttt{\_MT}$ & \textbf{61.45} & 1/8 & 1/4 & 1/2 \\
        %\hline
    \end{tabular}
    \caption{mIoU of \selene$_{\lambda_1=\lambda_2=0}$ and \selene$_{\lambda_2=0}$ obtained on validation set of PASCAL VOC for different fractions of the labeled training set.}
    \label{tab:ssl}
\end{table}

\begin{figure}[t!]
  \centering
 \begin{tabular}{cc}
\includegraphics[width=0.45\linewidth]{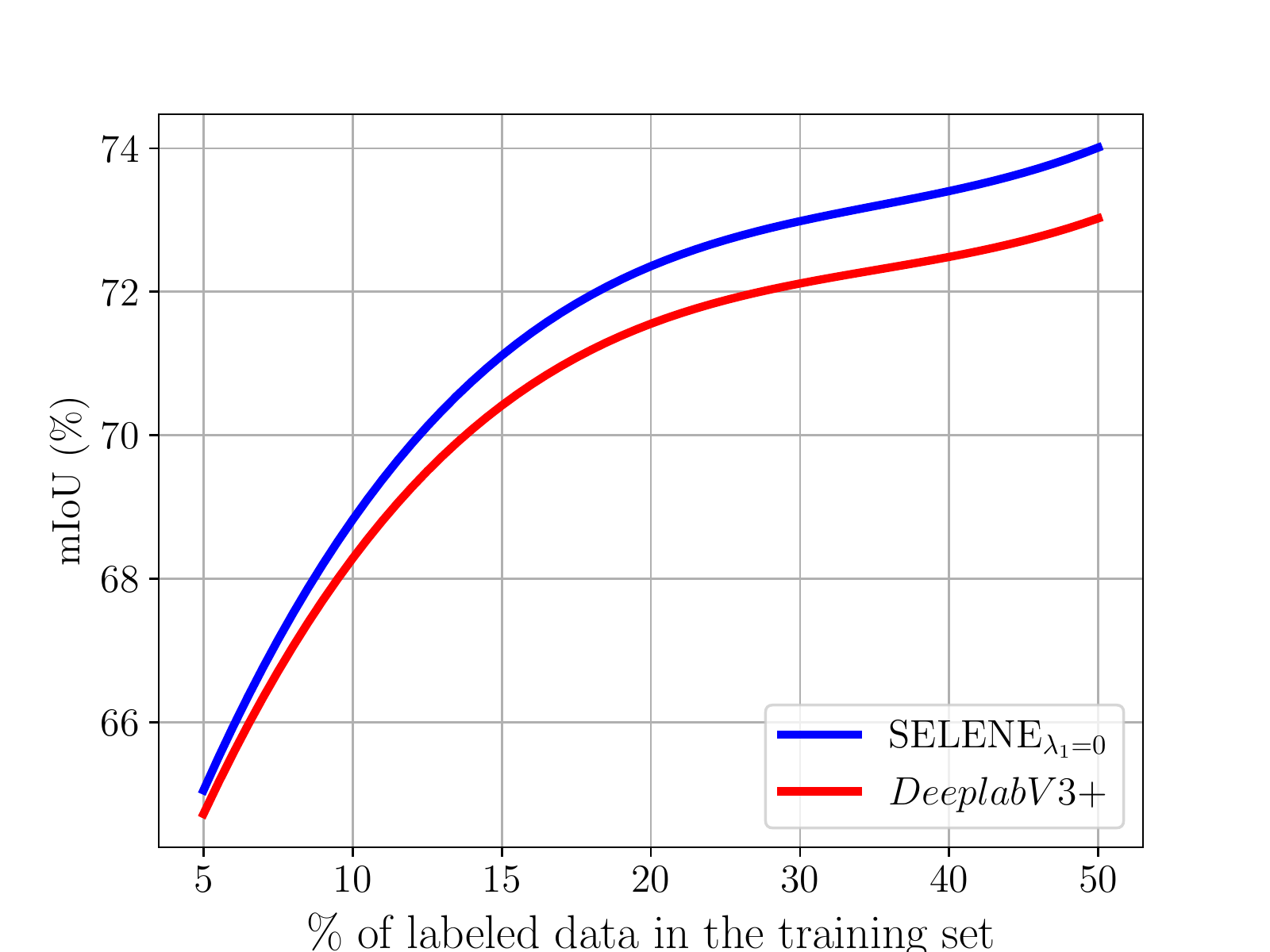} & \includegraphics[width=0.45\linewidth]{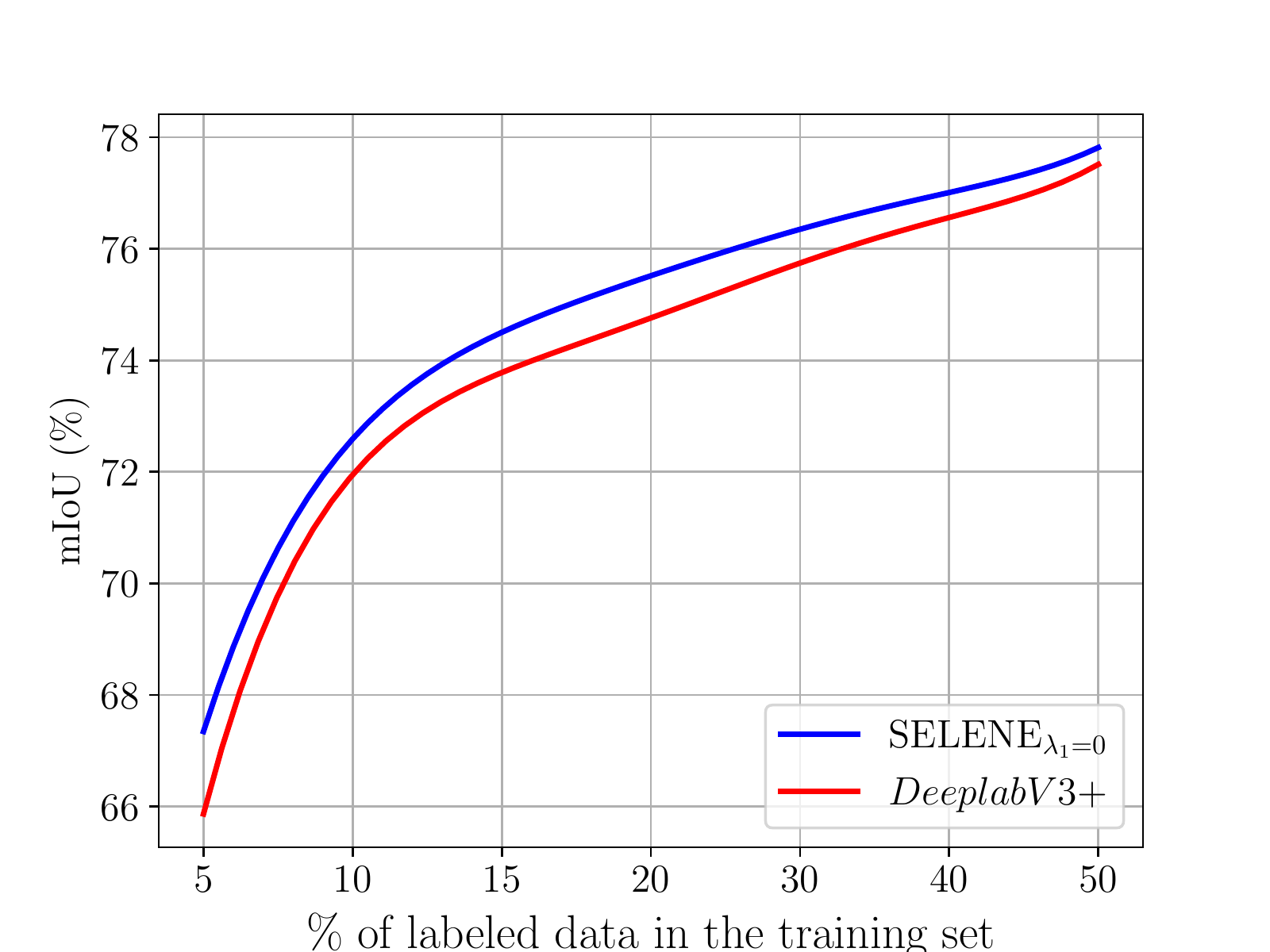}\\
(a) & (b) 
  \end{tabular}
  \caption{mIoU with respect to the percentage of labeled training set for the training of  DeepLabV3+ \cite{ChenZPSA18}, and \selene$_{\lambda_1=0}$ with the Mean-Teacher approach on Pascal VOC (a) and Cityscapes (b) datasets.}
  \label{fig:short}
\end{figure}
\subsubsection{Semi-supervised learning}
\renewcommand{\arraystretch}{1}
\begin{table}[t!]
    \centering
    \begin{tabular}{c c c c c}
\cmidrule[\heavyrulewidth](lr){2-5}
         \multicolumn{1}{l}{}  & \multicolumn{4}{c}{Fraction of the labeled training set} \\
        \cmidrule(lr){2-5}

          \multicolumn{1}{l}{} & $\frac{1}{16}$  & $\frac{1}{8}$  & $\frac{1}{4}$  & $\frac{1}{2}$   \\
\cmidrule[\heavyrulewidth](lr){2-5}
        \multicolumn{5}{l}{\textit{Mean teacher}}\\
        \toprule
         \selene$_{\lambda_1=0}$& \textbf{67.54} & \textbf{72.60} & \textbf{75.28} & \textbf{77.74}  \\
         DeeplabV3+ & 66.14 & 72.03 & 74.47 & 77.43  \\
        \bottomrule
        \multicolumn{5}{l}{\textit{Co-teaching}}\\
        \toprule
        \selene$_{\lambda_1=0}$ &  68.30 &  73.10 &  75.58 & 78.30\\
         DeeplabV3+ & \textbf{69.79} & \textbf{74.39} & \textbf{76.85} & \textbf{78.64}  \\
        \bottomrule
%        \textit{Co-teaching + CutMix}& $\selene{}^\dagger$ & 68.30 &  \\
%        \textit{}& Deeplabv3+ & 69.79 &  \\
%        \hline
    \end{tabular}
    \caption{mIoU of \selene$_{\lambda_1=0}$ and DeeplabV3+  \cite{ChenYZ021} obtained on Cityscapes validation set, using different fractions of the labeled training set with Mean-teacher (top) and co-training (down) approaches. Best results are shown in bold.}
    \label{tab:cityscape}
\end{table}

\begin{table}[t!]
    \centering
    \begin{tabular}{c c c c c}
\cmidrule[\heavyrulewidth](lr){2-5}
         \multicolumn{1}{l}{}  & \multicolumn{4}{c}{Fraction of the labeled training set} \\
        \cmidrule(lr){2-5}

          \multicolumn{1}{l}{} & $\frac{1}{16}$  & $\frac{1}{8}$  & $\frac{1}{4}$  & $\frac{1}{2}$   \\
\cmidrule[\heavyrulewidth](lr){2-5}
        \multicolumn{5}{l}{\textit{Mean teacher}}\\
         \toprule

        \selene$_{\lambda_1=0}$ & \textbf{65.05} & \textbf{68.64} & \textbf{72.19} & \textbf{73.95}  \\
        DeeplabV3+  & 64.72 & 68.12 & 71.41 & 72.97  \\
        
%        \textit{}& \textit{Deeplabv3+\cite{ChenYZ021}} & \textit{66.14} & \textit{72.03} & \textit{74.47} & \textit{77.43}  \\
       \bottomrule
        \multicolumn{5}{l}{\textit{Co-teaching}}\\
         \toprule
       \selene$_{\lambda_1=0}$ & 60.67 & 64.02 & 69.91 & 71.23 \\
        DeeplabV3+ & \textbf{64.13} & \textbf{69.52} & \textbf{71.77} & \textbf{74.11} \\
        \bottomrule
%        \textit{Co-teaching + CutMix}& $\selene{}^\dagger$ & 68.30 &  \\
%        \textit{}& Deeplabv3+ & 69.79 &  \\
%        \hline
    \end{tabular}
    \caption{mIoU of \selene$_{\lambda_1=0}$ and DeeplabV3+ obtained on Pascal VOC validation set. Best results are shown in bold.}
    \label{tab:pascal}
\end{table}
We now investigate the effect of semi-supervised learning alone by setting $\lambda_1$ to $0$. The resulting model is referred to as  \selene$_{\lambda_1=0}$ in the following. By setting $\lambda_1$ to $0$, we hence disable the associated geometric transformation. As semi-supervised techniques, we employed  Mean Teacher (Section \ref{sec:MT}) and Co-teaching (Section \ref{sec:CT}) techniques with the aim of analyzing the outcome of their underlying assumption in the behavior of \selene. For the Mean teacher method, the hyperparameter $\lambda_2$ \eqref{eq:totalloss} was empirically set to 100; and in the case of co-teaching, $\lambda_2$ was set to 5 for Cityscapes and in a range between 0.5 to 1.5 for Pascal VOC. Tables \ref{tab:cityscape} and \ref{tab:pascal} show the results for different fractions of the labeled training sets of the Cityscapes and Pascal VOC datasets, respectively. For comparison, we report the performance of DeepLabV3+ \cite{ChenZPSA18} based on ResNet-50, which employs a hand-crafted encoder-decoder structure. On both datasets, \selene$_{\lambda_1=0}$ performs better than DeeplabV3+ with the Mean Teacher strategy for all fractions of the labeled training set. We believe that this is because, with the Co-Training approach, the noise introduced when each of the classifiers assigns pseudo-labels to unlabeled examples has a snowball effect, reinforcing the predictions of the models in these errors and leading to an erroneous final model. Some approaches considered noise-label modelling jointly with the prediction function \cite{Krithara:08}. On the other hand, Mean Teacher is based on the consistency regularization with the only constraint that the outputs of the student and the teacher networks on the same unlabeled examples should be as close as possible. In this  case, there is no label-noise propagation and, with this approach, \selene{} is able to leverage the lack of label information by exploiting more efficiently the structure of the data using the unlabeled training set. These results are consistent with those of Figure \ref{fig:short} which depicts the evolution of mIoU with respect to the percentage of labeled training set for the training of DeepLabV3+ \cite{ChenZPSA18}, and \selene$_{\lambda_1=0}$ on Pascal VOC and Cityscapes using the Co-Teaching technique.

% \begin{figure}
% \centering
% \begin{tikzpicture}
% \begin{axis}[
%     xlabel={Fraction of supervised samples},
%     ylabel={mIoU (\%)},
%     xticklabels={$\frac{1}{16}$,$\frac{1}{8}$,$\frac{1}{4}$,$\frac{1}{2}$},
%     %x tick label style={rotate=45,anchor=east},
%     %xticklabels={$\frac{\pgfmathprintnumber{\tick}}{16}$}
%     xtick={1,2,3,4},
%     %ytick={60,65,70,75,80},
%     legend pos=north west,
%     ymajorgrids=true,
%     grid style=dashed,
% ]

% \addplot[
%     color=blue,
%     mark=square,
%     ]
%     coordinates {
%     (1,67.54)(2,72.60)(3,75.28)(4,77.74)
%     };
% \addplot[
%     color=red,
%     mark=x,
%     ]
%     coordinates {
%     (1,66.14)(2,72.03)(3,74.47)(4,77.43)
%     };
% \legend{$\selene{}\dagger$,DeeplabV3+ }
% \end{axis}
% \end{tikzpicture}
% \caption{Results on Cityscapes val. set, under various partition, using Mean Teacher method. As we follow the same training protocol, the  plotted DeeplabV3+ results are from \cite{ChenYZ021}.}
% \label{fig:mt_city}
% \end{figure}

\renewcommand{\arraystretch}{1}
\begin{table}[t!]
    \centering
    \begin{tabular}{ccccc}
 \cmidrule[\heavyrulewidth](lr){2-5}
         \multicolumn{1}{l}{}  & \multicolumn{4}{c}{Fraction of the labeled training set} \\
        \cmidrule(lr){2-5}

          \multicolumn{1}{l}{} &$\frac{1}{16}$ & $\frac{1}{8}$ & $\frac{1}{4}$  & $\frac{1}{2}$  \\        \toprule
       \scriptsize \selene{}$_{\lambda_1=\lambda_2=0}$ & 53.33 & 59.45 & 65.21 & 69.02\\
%       \hline
       \selene{}$_{\lambda_2=0}$ & 53.80 & 60.33 & 65.25 & 69.27 \\
 %       \hline
    \selene{}$_{\lambda_1=0}$ & 65.05 & 68.64 & 72.19 & 73.95  \\
  %  \hline
    \selene{} & \textbf{65.32} & \textbf{68.96} & \textbf{72.49} & \textbf{74.47} \\
\bottomrule
%        \textit{Co-teaching + CutMix}& $\selene{}^\dagger$ & 68.30 &  \\
%        \textit{}& Deeplabv3+ & 69.79 &  \\
%        \hline
    \end{tabular}
    \caption{mIoU of \selene{} and its variants obtained on the validation  set of PASCAL VOC.}
    \label{tab:summary}
\end{table}

\begin{figure}[b!]
\centering
\begin{tikzpicture}[thick, scale=0.95]
\begin{axis}[
    xtick={1,2},
    xticklabels={Cityscapes ,Pascal VOC},
    %x tick label style={rotate=0,anchor=east},
    ylabel=FLOPs(G),
    ybar,
    ymajorgrids=true,
    ymin=0, ymax=600,
    ytick={50,200,400,600},
    enlarge x limits={abs = .5},
    enlarge y limits={value=0.15, upper},
    bar width=0.4,
    width=0.92\linewidth,
    nodes near coords,
    %every node near coord/.append style={font=\footnotesize},
]
\addplot 
	coordinates {(1,479) (2,62) 
		 };
\addplot 
	coordinates {(1,120)  (2,15) };
\legend{DeeplabV3+, \selene{}}
\end{axis}
\end{tikzpicture}
\caption{Comparison in terms of FLOPs(G) between  \selene{} and DeeplabV3+ on Pascal VOC and Cityscapes datasets.}
\label{fig:flops}
\end{figure}
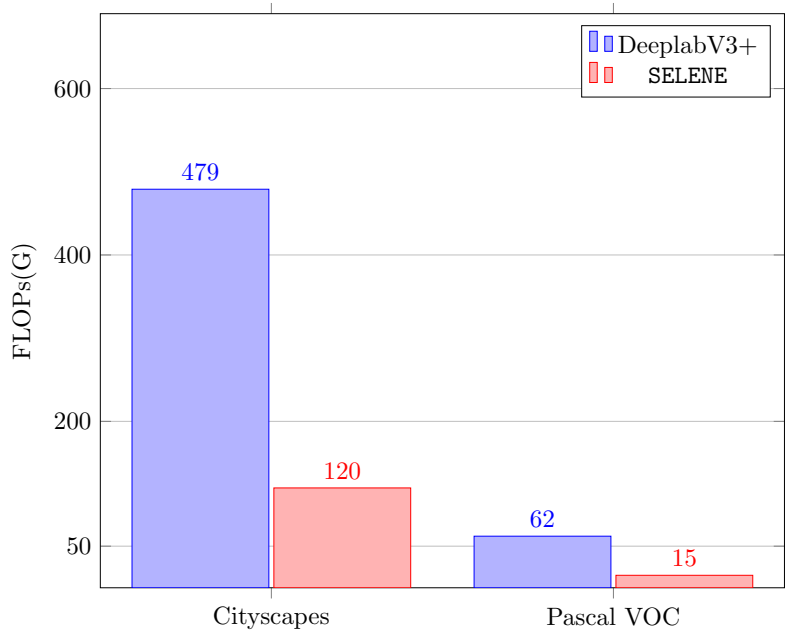 
\subsubsection{Self-supervised and Semi-supervised learning}

Finally, in Table \ref{tab:summary} we compare the performance of \selene{} with self-supervised and semi-supervised learning using the Mean-Teacher technique to its other versions discussed above on the Pascal VOC dataset. The pretext task has also here a limited benefit on semi-supervised learning, comforting the idea that, while complementing, puzzle solving and pixel classification are not totally correlated. In Figure \ref{fig:flops}, we compare DeeplabV3+ and \selene{} with the co-teaching approach in terms of floating-point operations. It comes out that \selene{} use up to 4 times less floating-point operations with respect to this measure than the neural-network with the hand-crafted architecture.

\section{Summary \& Outlook}
\label{conclu}
In this study, we proposed a self-supervised semi-supervised learning approach with Neural Architecture Search for semantic segmentation. We showed on Pascal VOC and Cityscapes datasets that by jointly solving a jigsaw pretext task discovered using self-learning over unlabeled training data and leveraging the structure of the unlabeled data with Co-Teaching, our approach finds an efficient neural network  model for this task.  
Concerning the limitations of our study, we can include the fact that we study only a single self-supervised task, which could not relate the full potential of the self-supervised regularization. Another point, could be the proposed framework is sensitive to the choice of techniques used, which are not all equally efficient, as well as the number of data point used which in some cases can pose performance problems. This work opens up new perspectives, including the hyperparameters tuning with some dedicated techniques \cite{FalknerKH18}\cite{LiJDRT17}.  Moreover, we found that combining self-supervised and semi-supervised learning is promising, and that alternative self-supervised learning strategies, such as rotation \cite{GidarisSK18} or the more recent contrastive technique \cite{He0WXG20,CaronMMGBJ20} should be investigated for semantic segmentation. Another point could also be the potential improvement of the decoder by using a spatial pooling module (e.g. PSP\cite{ZhaoSQWJ17}, ASPP\cite{ChenPKMY18}), as in \cite{LiSCLZWS20}. 
%In a more technical way, we could consider an optimization of the training process by using for example the asynchronous distributed training or the precision reduction in FP16. 

%-------------------------------------------------------------------------

\balance
%%%%%%%%% REFERENCES
{\small
\bibliographystyle{ieee_fullname}
\bibliography{egbib}

\begin{thebibliography}{10}\itemsep=-1pt

\bibitem{AbbasnejadDH17}
M.~Ehsan Abbasnejad, Anthony~R. Dick, and Anton van~den Hengel.
\newblock Infinite variational autoencoder for semi-supervised learning.
\newblock In {\em CVPR}, pages 781--790. {IEEE} Computer Society, 2017.

\bibitem{Amini:2008}
Massih{-}Reza Amini, Fran{\c{c}}ois Laviolette, and Nicolas Usunier.
\newblock A transductive bound for the voted classifier with an application to
  semi-supervised learning.
\newblock In {\em Advances in Neural Information Processing Systems}, pages
  65--72, 2008.

\bibitem{AminiU15}
Massih{-}Reza Amini and Nicolas Usunier.
\newblock {\em Learning with Partially Labeled and Interdependent Data}.
\newblock Springer, 2015.

\bibitem{AthiwaratkunFIW19}
Ben Athiwaratkun, Marc Finzi, Pavel Izmailov, and Andrew~Gordon Wilson.
\newblock There are many consistent explanations of unlabeled data: Why you
  should average.
\newblock In {\em ICLR}. OpenReview.net, 2019.

\bibitem{badrinarayanan2017segnet}
Vijay Badrinarayanan, Alex Kendall, and Roberto Cipolla.
\newblock Segnet: A deep convolutional encoder-decoder architecture for image
  segmentation.
\newblock {\em IEEE transactions on pattern analysis and machine intelligence},
  39(12):2481--2495, 2017.

\bibitem{remixmatch}
David Berthelot, Nicholas Carlini, Ekin~D. Cubuk, Alex Kurakin, Kihyuk Sohn,
  Han Zhang, and Colin Raffel.
\newblock Remixmatch: Semi-supervised learning with distribution alignment and
  augmentation anchoring.
\newblock {\em CoRR}, abs/1911.09785, 2019.

\bibitem{BerthelotCGPOR19}
David Berthelot, Nicholas Carlini, Ian~J. Goodfellow, Nicolas Papernot, Avital
  Oliver, and Colin Raffel.
\newblock Mixmatch: {A} holistic approach to semi-supervised learning.
\newblock In Hanna~M. Wallach, Hugo Larochelle, Alina Beygelzimer, Florence
  d'Alch{\'{e}}{-}Buc, Emily~B. Fox, and Roman Garnett, editors, {\em NeurIPS},
  pages 5050--5060, 2019.

\bibitem{BeyerZOK19}
Lucas Beyer, Xiaohua Zhai, Avital Oliver, and Alexander Kolesnikov.
\newblock {S4L:} self-supervised semi-supervised learning.
\newblock In {\em ICCV}, pages 1476--1485. {IEEE}, 2019.

\bibitem{CaronMMGBJ20}
Mathilde Caron, Ishan Misra, Julien Mairal, Priya Goyal, Piotr Bojanowski, and
  Armand Joulin.
\newblock Unsupervised learning of visual features by contrasting cluster
  assignments.
\newblock In {\em NeurIPS}, 2020.

\bibitem{Chap06}
Olivier Chapelle, Bernhard Schölkopf, and Alexander Zien, editors.
\newblock {\em Semi-Supervised Learning}.
\newblock The MIT Press, 2006.

\bibitem{ChenCZPZSAS18}
Liang{-}Chieh Chen, Maxwell~D. Collins, Yukun Zhu, George Papandreou, Barret
  Zoph, Florian Schroff, Hartwig Adam, and Jonathon Shlens.
\newblock Searching for efficient multi-scale architectures for dense image
  prediction.
\newblock In {\em NeurIPS}, pages 8713--8724, 2018.

\bibitem{ChenPKMY18}
Liang{-}Chieh Chen, George Papandreou, Iasonas Kokkinos, Kevin Murphy, and
  Alan~L. Yuille.
\newblock Deeplab: Semantic image segmentation with deep convolutional nets,
  atrous convolution, and fully connected crfs.
\newblock {\em {IEEE} Trans. Pattern Anal. Mach. Intell.}, 40(4):834--848,
  2018.

\bibitem{ChenZPSA18}
Liang{-}Chieh Chen, Yukun Zhu, George Papandreou, Florian Schroff, and Hartwig
  Adam.
\newblock Encoder-decoder with atrous separable convolution for semantic image
  segmentation.
\newblock In {\em European Conference in Computer Vision {ECCV}}, volume 11211,
  pages 833--851. Springer, 2018.

\bibitem{ChenYZ021}
Xiaokang Chen, Yuhui Yuan, Gang Zeng, and Jingdong Wang.
\newblock Semi-supervised semantic segmentation with cross pseudo supervision.
\newblock In {\em CVPR}, pages 2613--2622. Computer Vision Foundation / {IEEE},
  2021.

\bibitem{CordtsORREBFRS16}
Marius Cordts, Mohamed Omran, Sebastian Ramos, Timo Rehfeld, Markus Enzweiler,
  Rodrigo Benenson, Uwe Franke, Stefan Roth, and Bernt Schiele.
\newblock The cityscapes dataset for semantic urban scene understanding.
\newblock In {\em CVPR}, pages 3213--3223. {IEEE} Computer Society, 2016.

\bibitem{DaiYYCS17}
Zihang Dai, Zhilin Yang, Fan Yang, William~W. Cohen, and Ruslan Salakhutdinov.
\newblock Good semi-supervised learning that requires a bad {GAN}.
\newblock In Isabelle Guyon, Ulrike von Luxburg, Samy Bengio, Hanna~M. Wallach,
  Rob Fergus, S.~V.~N. Vishwanathan, and Roman Garnett, editors, {\em NIPS},
  pages 6510--6520, 2017.

\bibitem{DoerschGE15}
Carl Doersch, Abhinav Gupta, and Alexei~A. Efros.
\newblock Unsupervised visual representation learning by context prediction.
\newblock In {\em 2015 {IEEE} International Conference on Computer Vision,
  {ICCV}}, pages 1422--1430. {IEEE} Computer Society, 2015.

\bibitem{DosovitskiyFSRB16}
Alexey Dosovitskiy, Philipp Fischer, Jost~Tobias Springenberg, Martin~A.
  Riedmiller, and Thomas Brox.
\newblock Discriminative unsupervised feature learning with exemplar
  convolutional neural networks.
\newblock {\em {IEEE} Transactions in Pattern Analysis and Machine
  Intelligence}, 38(9):1734--1747, 2016.

\bibitem{EveringhamGWWZ10}
Mark Everingham, Luc~Van Gool, Christopher K.~I. Williams, John~M. Winn, and
  Andrew Zisserman.
\newblock The pascal visual object classes {(VOC)} challenge.
\newblock {\em Int. J. Comput. Vis.}, 88(2):303--338, 2010.

\bibitem{FalknerKH18}
Stefan Falkner, Aaron Klein, and Frank Hutter.
\newblock {BOHB:} robust and efficient hyperparameter optimization at scale.
\newblock In Jennifer~G. Dy and Andreas Krause, editors, {\em ICML}, volume~80
  of {\em Proceedings of Machine Learning Research}, pages 1436--1445. {PMLR},
  2018.

\bibitem{FourureEFMT017}
Damien Fourure, R{\'{e}}mi Emonet, {\'{E}}lisa Fromont, Damien Muselet, Alain
  Tr{\'{e}}meau, and Christian Wolf.
\newblock Residual conv-deconv grid network for semantic segmentation.
\newblock In {\em BMVC}, 2017.

\bibitem{french2019semi}
Geoffrey French, Samuli Laine, Timo Aila, Michal Mackiewicz, and Graham~D.
  Finlayson.
\newblock Semi-supervised semantic segmentation needs strong, varied
  perturbations.
\newblock In {\em 31st British Machine Vision Conference 2020, {BMVC}}. {BMVA}
  Press, 2020.

\bibitem{GidarisSK18}
Spyros Gidaris, Praveer Singh, and Nikos Komodakis.
\newblock Unsupervised representation learning by predicting image rotations.
\newblock In {\em ICLR}, 2018.

\bibitem{GrillSATRBDPGAP20}
Jean{-}Bastien Grill, Florian Strub, Florent Altch{\'{e}}, Corentin Tallec,
  Pierre~H. Richemond, Elena Buchatskaya, Carl Doersch, Bernardo~{\'{A}}vila
  Pires, Zhaohan Guo, Mohammad~Gheshlaghi Azar, Bilal Piot, Koray Kavukcuoglu,
  R{\'{e}}mi Munos, and Michal Valko.
\newblock Bootstrap your own latent - {A} new approach to self-supervised
  learning.
\newblock In {\em NeurIPS}, 2020.

\bibitem{HariharanABMM11}
Bharath Hariharan, Pablo Arbelaez, Lubomir~D. Bourdev, Subhransu Maji, and
  Jitendra Malik.
\newblock Semantic contours from inverse detectors.
\newblock In {\em ICCV}, pages 991--998. {IEEE} Computer Society, 2011.

\bibitem{He0WXG20}
Kaiming He, Haoqi Fan, Yuxin Wu, Saining Xie, and Ross~B. Girshick.
\newblock Momentum contrast for unsupervised visual representation learning.
\newblock In {\em CVPR}, pages 9726--9735. Computer Vision Foundation / {IEEE},
  2020.

\bibitem{HeZRS15}
Kaiming He, Xiangyu Zhang, Shaoqing Ren, and Jian Sun.
\newblock Delving deep into rectifiers: Surpassing human-level performance on
  imagenet classification.
\newblock In {\em ICCV}, pages 1026--1034. {IEEE} Computer Society, 2015.

\bibitem{JingT21}
Longlong Jing and Yingli Tian.
\newblock Self-supervised visual feature learning with deep neural networks:
  {A} survey.
\newblock {\em {IEEE} Trans. Pattern Anal. Mach. Intell.}, 43(11):4037--4058,
  2021.

\bibitem{KeWYRL19}
Zhanghan Ke, Daoye Wang, Qiong Yan, Jimmy S.~J. Ren, and Rynson W.~H. Lau.
\newblock Dual student: Breaking the limits of the teacher in semi-supervised
  learning.
\newblock In {\em ICCV}, pages 6727--6735. {IEEE}, 2019.

\bibitem{KingmaMRW14}
Diederik~P. Kingma, Shakir Mohamed, Danilo~Jimenez Rezende, and Max Welling.
\newblock Semi-supervised learning with deep generative models.
\newblock In {\em NeurIPS}, pages 3581--3589, 2014.

\bibitem{Krithara:08}
Anastasia Krithara, Massih-Reza Amini, Jean-Michel Renders, and Cyril Goutte.
\newblock Semi-supervised document classification with a mislabeling error
  model.
\newblock In {\em 30th European Conference on Information Retrieval (ECIR'05)},
  pages 370--381, Glasgow, 2008.

\bibitem{LaineA17}
Samuli Laine and Timo Aila.
\newblock Temporal ensembling for semi-supervised learning.
\newblock In {\em ICLR}. OpenReview.net, 2017.

\bibitem{LiJDRT17}
Lisha Li, Kevin~G. Jamieson, Giulia DeSalvo, Afshin Rostamizadeh, and Ameet
  Talwalkar.
\newblock Hyperband: {A} novel bandit-based approach to hyperparameter
  optimization.
\newblock {\em J. Mach. Learn. Res.}, 18:185:1--185:52, 2017.

\bibitem{LiSCLZWS20}
Yanwei Li, Lin Song, Yukang Chen, Zeming Li, Xiangyu Zhang, Xingang Wang, and
  Jian Sun.
\newblock Learning dynamic routing for semantic segmentation.
\newblock In {\em CVPR}, pages 8550--8559. Computer Vision Foundation / {IEEE},
  2020.

\bibitem{LiuDHGYX20}
Chenxi Liu, Piotr Doll{\'{a}}r, Kaiming He, Ross~B. Girshick, Alan~L. Yuille,
  and Saining Xie.
\newblock Are labels necessary for neural architecture search?
\newblock In {\em ECCV}, pages 798--813, 2020.

\bibitem{LiuDY19}
Xiaolong Liu, Zhidong Deng, and Yuhan Yang.
\newblock Recent progress in semantic image segmentation.
\newblock {\em Artif. Intell. Rev.}, 52(2):1089--1106, 2019.

\bibitem{LongSD15}
Jonathan Long, Evan Shelhamer, and Trevor Darrell.
\newblock Fully convolutional networks for semantic segmentation.
\newblock In {\em CVPR}, pages 3431--3440. {IEEE} Computer Society, 2015.

\bibitem{MiyatoMKI19}
Takeru Miyato, Shin{-}ichi Maeda, Masanori Koyama, and Shin Ishii.
\newblock Virtual adversarial training: {A} regularization method for
  supervised and semi-supervised learning.
\newblock {\em {IEEE} Transactions in Pattern Analysis and Machine
  Intelligence}, 41(8):1979--1993, 2019.

\bibitem{Mottaghi14}
Roozbeh Mottaghi, Xianjie Chen, Xiaobai Liu, Nam-Gyu Cho, Seong-Whan Lee, Sanja
  Fidler, Raquel Urtasun, and Alan Yuille.
\newblock The role of context for object detection and semantic segmentation in
  the wild.
\newblock In {\em 2014 IEEE Conference on Computer Vision and Pattern
  Recognition}, pages 891--898, 2014.

\bibitem{muller2021convolutional}
David M{\"u}ller, Andreas Ehlen, and Bernd Valeske.
\newblock Convolutional neural networks for semantic segmentation as a tool for
  multiclass face analysis in thermal infrared.
\newblock {\em Journal of nondestructive evaluation}, 40(1):1--10, 2021.

\bibitem{NekrasovCS019}
Vladimir Nekrasov, Hao Chen, Chunhua Shen, and Ian~D. Reid.
\newblock Fast neural architecture search of compact semantic segmentation
  models via auxiliary cells.
\newblock In {\em CVPR}, pages 9126--9135. Computer Vision Foundation / {IEEE},
  2019.

\bibitem{NorooziF16}
Mehdi Noroozi and Paolo Favaro.
\newblock Unsupervised learning of visual representations by solving jigsaw
  puzzles.
\newblock In {\em ECCV}, volume 9910 of {\em Lecture Notes in Computer
  Science}, pages 69--84. Springer, 2016.

\bibitem{Odena16a}
Augustus Odena.
\newblock Semi-supervised learning with generative adversarial networks.
\newblock {\em CoRR}, abs/1606.01583, 2016.

\bibitem{OlssonTPS21}
Viktor Olsson, Wilhelm Tranheden, Juliano Pinto, and Lennart Svensson.
\newblock Classmix: Segmentation-based data augmentation for semi-supervised
  learning.
\newblock In {\em {IEEE} Winter Conference on Applications of Computer Vision,
  {WACV} 2021, Waikoloa, HI, USA, January 3-8, 2021}, pages 1368--1377. {IEEE},
  2021.

\bibitem{Ouali2020AnOO}
Yassine Ouali, C{\'e}line Hudelot, and Myriam Tami.
\newblock An overview of deep semi-supervised learning.
\newblock {\em ArXiv}, abs/2006.05278, 2020.

\bibitem{OualiHT20}
Yassine Ouali, C{\'{e}}line Hudelot, and Myriam Tami.
\newblock Semi-supervised semantic segmentation with cross-consistency
  training.
\newblock In {\em CVPR}, pages 12671--12681. Computer Vision Foundation /
  {IEEE}, 2020.

\bibitem{pytorch}
Adam Paszke, Sam Gross, Francisco Massa, Adam Lerer, James Bradbury, Gregory
  Chanan, Trevor Killeen, Zeming Lin, Natalia Gimelshein, Luca Antiga, Alban
  Desmaison, Andreas Kopf, Edward Yang, Zachary DeVito, Martin Raison, Alykhan
  Tejani, Sasank Chilamkurthy, Benoit Steiner, Lu Fang, Junjie Bai, and Soumith
  Chintala.
\newblock Pytorch: An imperative style, high-performance deep learning library.
\newblock In {\em NeurIPS}, pages 8024--8035. 2019.

\bibitem{PathakKDDE16}
Deepak Pathak, Philipp Kr{\"{a}}henb{\"{u}}hl, Jeff Donahue, Trevor Darrell,
  and Alexei~A. Efros.
\newblock Context encoders: Feature learning by inpainting.
\newblock In {\em CVPR}, pages 2536--2544. {IEEE} Computer Society, 2016.

\bibitem{QiaoSZWY18}
Siyuan Qiao, Wei Shen, Zhishuai Zhang, Bo Wang, and Alan~L. Yuille.
\newblock Deep co-training for semi-supervised image recognition.
\newblock In Vittorio Ferrari, Martial Hebert, Cristian Sminchisescu, and Yair
  Weiss, editors, {\em ECCV}, volume 11219 of {\em Lecture Notes in Computer
  Science}, pages 142--159. Springer, 2018.

\bibitem{RasmusBHVR15}
Antti Rasmus, Mathias Berglund, Mikko Honkala, Harri Valpola, and Tapani Raiko.
\newblock Semi-supervised learning with ladder networks.
\newblock In {\em NeurIPS}, pages 3546--3554, 2015.

\bibitem{RezatofighiTGS019}
Hamid Rezatofighi, Nathan Tsoi, JunYoung Gwak, Amir Sadeghian, Ian~D. Reid, and
  Silvio Savarese.
\newblock Generalized intersection over union: {A} metric and a loss for
  bounding box regression.
\newblock In {\em Conference on Computer Vision and Pattern Recognition,
  {CVPR}}, pages 658--666, 2019.

\bibitem{ronneberger2015u}
Olaf Ronneberger, Philipp Fischer, and Thomas Brox.
\newblock U-net: Convolutional networks for biomedical image segmentation.
\newblock In {\em International Conference on Medical image computing and
  computer-assisted intervention}, pages 234--241. Springer, 2015.

\bibitem{ShrivastavaGG16}
Abhinav Shrivastava, Abhinav Gupta, and Ross~B. Girshick.
\newblock Training region-based object detectors with online hard example
  mining.
\newblock In {\em CVPR}, pages 761--769. {IEEE} Computer Society, 2016.

\bibitem{SohnBCZZRCKL20}
Kihyuk Sohn, David Berthelot, Nicholas Carlini, Zizhao Zhang, Han Zhang, Colin
  Raffel, Ekin~Dogus Cubuk, Alexey Kurakin, and Chun{-}Liang Li.
\newblock Fixmatch: Simplifying semi-supervised learning with consistency and
  confidence.
\newblock In Hugo Larochelle, Marc'Aurelio Ranzato, Raia Hadsell,
  Maria{-}Florina Balcan, and Hsuan{-}Tien Lin, editors, {\em NeurIPS}, 2020.

\bibitem{TarvainenV17}
Antti Tarvainen and Harri Valpola.
\newblock Mean teachers are better role models: Weight-averaged consistency
  targets improve semi-supervised deep learning results.
\newblock In {\em NeurIPS}, pages 1195--1204, 2017.

\bibitem{EngelenH20}
Jesper~E. van Engelen and Holger~H. Hoos.
\newblock A survey on semi-supervised learning.
\newblock {\em Mach. Learn.}, 109(2):373--440, 2020.

\bibitem{Wu2020reid}
Xifang Wu, Songlin Sun, and Meixia Fu.
\newblock Person re-identification based on semantic segmentation.
\newblock In Yue Wang, Meixia Fu, Lexi Xu, and Jiaqi Zou, editors, {\em Signal
  and Information Processing, Networking and Computers}, 2020.

\bibitem{XieLHL20}
Qizhe Xie, Minh{-}Thang Luong, Eduard~H. Hovy, and Quoc~V. Le.
\newblock Self-training with noisy student improves imagenet classification.
\newblock In {\em CVPR}, pages 10684--10695, 2020.

\bibitem{YuWPGYS18}
Changqian Yu, Jingbo Wang, Chao Peng, Changxin Gao, Gang Yu, and Nong Sang.
\newblock Bisenet: Bilateral segmentation network for real-time semantic
  segmentation.
\newblock In {\em ECCV}, volume 11217 of {\em Lecture Notes in Computer
  Science}, pages 334--349. Springer, 2018.

\bibitem{ZhangIE16}
Richard Zhang, Phillip Isola, and Alexei~A. Efros.
\newblock Colorful image colorization.
\newblock In {\em European Conference in Computer Vision {ECCV}}, pages
  649--666, 2016.

\bibitem{ZhaoSQWJ17}
Hengshuang Zhao, Jianping Shi, Xiaojuan Qi, Xiaogang Wang, and Jiaya Jia.
\newblock Pyramid scene parsing network.
\newblock In {\em CVPR}, pages 6230--6239. {IEEE} Computer Society, 2017.

\bibitem{zhu2005semi}
Xiaojin~Jerry Zhu.
\newblock Semi-supervised learning literature survey.
\newblock 2005.

\end{thebibliography}
}

\end{document}